\documentclass[11pt]{article}
\usepackage[T1]{fontenc}
\usepackage[utf8]{inputenc}
\usepackage{amsfonts}
\usepackage{amsmath}
\usepackage{enumerate}
\usepackage{amssymb}
\usepackage{amsthm}
\usepackage{bbm}
\usepackage{bm}
\usepackage{mathrsfs}
\usepackage{verbatim}
\usepackage{setspace}
\usepackage{color}
\usepackage{pdfsync}
\usepackage{booktabs}
\usepackage{tikz-cd}

\usepackage{tikz}

\usepackage{graphicx}
\usepackage{amsmath,varwidth,array,ragged2e}
\usepackage{tabu}
\usepackage{array}
\usepackage{makecell}

\usepackage{enumitem}
\theoremstyle{plain}
\newtheorem{theorem}{Theorem}

\newtheorem{lemma}[theorem]{Lemma}

\theoremstyle{definition}

\theoremstyle{remark}

\renewenvironment{thebibliography}[1]{%
	\begin{oldthebibliography}{#1}%
		\setlength{\baselineskip}{.9em}
		\linespread{1}
		\small
		\setlength{\parskip}{0.3ex}%
		\setlength{\itemsep}{.5em}%
	}%
	{%
	\end{oldthebibliography}%
}

\DeclareMathOperator{\Lip}{Lip}

\numberwithin{equation}{section}

\usepackage[pdfborder={0 0 0}]{hyperref}
\hypersetup{
	urlcolor = black,
	pdfauthor = {},
	pdfkeywords = {T.B.D.;
	},
}

\title{Lipschitz neural networks are dense in the set of all Lipschitz functions}
\author{Stephan Eckstein\thanks{Department of Mathematics, University of 
Konstanz, Universit\"{a}tsstraße 10, 78464 Konstanz, Germany, 
stephan.eckstein@uni-konstanz.de}}

\begin{document}
	
\maketitle

\begin{abstract}
This note shows that, for a fixed Lipschitz constant $L > 0$, one layer neural 
networks that are $L$-Lipschitz are 
dense in the set of all $L$-Lipschitz functions with respect to the uniform
norm on bounded sets.
\end{abstract}

\section{Introduction and main result}
Let $d \in \mathbb{N}$, $K \subset \mathbb{R}^d$ be bounded and $L > 0$. We fix 
a norm 
$\|\cdot\|$ on $\mathbb{R}^d$ and for a function $f: \mathbb{R}^d \rightarrow 
\mathbb{R}$ we recall the uniform norm on $K$ given by $\|f\|_{\infty, K} = 
\sup_{x\in K}|f(x)|$. Let 
$\Lip_{L, K}$ be the set of 
all
functions 
mapping from $\mathbb{R}^d$ to $\mathbb{R}$ that are $L$-Lipschitz on $K$, 
i.e., 
all functions $f : 
\mathbb{R}^d \rightarrow \mathbb{R}$ satisfying 
$|f(x)-f(y)| \leq L\|x-y\|$ for all $x, y \in K$.

We further fix an activation function $\varphi : \mathbb{R} \rightarrow 
\mathbb{R}$ and define the set $N^m$ of all one layer neural 
networks with layer-width $m \in \mathbb{N}$ mapping 
$\mathbb{R}^d$ to $\mathbb{R}$, i.e., $f_m \in 
N^m$ can be written as
\[
f_m(x) = b + \sum_{i=1}^m a_i\, \varphi\Big(\sum_{j=1}^d w_{i, j} x_j + 
c_i\Big) 
~~ 
\text{ 
	for all } x \in \mathbb{R}^d,
\]
where $b, a_1, ..., a_m, w_{1, 1}, ..., w_{m, d}, c_1, ..., c_m \in \mathbb{R}$ 
are the parameters of the network $f_m$.

Approximation properties of the set $N^m$ are well studied (see, e.g., 
\cite{hornik1991approximation, pinkus1999approximation}). 
In this note however, we study approximation properties of the set 
$\Lip_{L, K}^m
:= 
\Lip_{L, K} \cap N^m$. We consider the question of 
approximating functions in $\Lip_{L, K}$ by networks in 
$\Lip_{L, K}^m$. Related questions were studied in \cite{anil2019sorting, 
	huster2018limitations} and working with neural networks under a Lipschitz 
constraint occurs in many problems related to Wasserstein distances (see, e.g., 
\cite{arjovsky2017wasserstein, ozair2019wasserstein}) and regularization and 
adversarial robustness (see, e.g., 
\cite{cisse2017parseval,gouk2018regularisation, tsuzuku2018lipschitz}). Even 
though in practice, enforcing a Lipschitz constraint for neural networks has to 
rely 
on either penalization methods (see, e.g., \cite{gulrajani2017improved, 
	petzka2018regularization}) or special architectures or weight restrictions 
(see, e.g., \cite{anil2019sorting,arjovsky2017wasserstein, 
	miyato2018spectral}), the set $\Lip_{L, K}^m$ can be regarded as an 
idealized 
version of working with neural networks under a Lipschitz constraint. This note 
shows, under mild assumptions on the activation function, that the addition of 
a Lipschitz constraint does not inhibit the expressiveness of neural networks.
The main result is the following:

\begin{theorem}
	\label{thm:main}
	Let $\varphi$ be one time continuously differentiable and not polynomial,
	or let $\varphi$ be the ReLU. Then it holds:
	
	For any $\varepsilon > 0$, there exists some $m = m(\varepsilon) \in 
	\mathbb{N}$ so that 
	\[
	\sup_{f \in \Lip_{L, K}} \inf_{f_m \in \Lip_{L, K}^m} \|f-f_m\|_{\infty, K} 
	\leq \varepsilon.
	\]
\end{theorem}

The proof of Theorem \ref{thm:main} relies on existing work on neural 
network 
approximations of functions and their derivatives. The references are 
\cite{pinkus1999approximation}
for the case of a continuously differentiable activation functions, and 
\cite{ito1994differentiable} for 
the ReLU. Instead of the ReLU, other weakly differentiable activation 
functions could be considered which satisfy the assumptions of \cite[Theorem 
4.1, 4.2 or 4.3]{ito1994differentiable}.

The usual methods apply when transitioning from shallow networks (with 
one hidden layer) to many-layer networks. The result still holds, since the 
later 
layers can approximate the identity function under a Lipschitz constraint up to 
arbitrary accuracy.


\section{Proof of Theorem \ref{thm:main}}
For the proof of Theorem 1, we will first show in Subsection 
\ref{subsec:proofpart1} that a simpler statement holds, where the size 
$m=m(\varepsilon, f)$ of the network may depend on the Lipschitz function $f 
\in \Lip_{L, K}$ to be approximated. The general case is a simple consequence 
and is shown in 
Subsection \ref{subsec:proofpart2}. First, we state simplifications 
which will be used in the first part of the proof.
\subsection{Scaling and simplifications}\label{subsec:scaling} We only show the 
statements for $L=1$. This may be done since neural networks can be multiplied 
by a constant. Thus, instead of approximating $f\in \Lip_{L, K}$ up to accuracy 
$\varepsilon$, one may approximate the function $\frac{f}{L}$ up to accuracy 
$\frac{\varepsilon}{L}$ and then scale the approximating networks by the factor 
$L$. 

Analogously to the Lipschitz constant, we assume that the considered norms are 
normalized to $\max_{x \in [0, 1]^d}\|x\| = 1$, which means in particular that 
$\|x\|_1 = \frac{1}{d} 
\sum_{i=1}^d |x_i|$.

We also assume that any function to be approximated is normalized to $f(0) = 
0$. This is not a restriction, since neural networks can be shifted by 
constants, and hence one can first approximate the function $f - f(0)$ and then 
shift the neural network by the constant $f(0)$.

Further, we assume without loss of generality that a function $f \in \Lip_{1, 
	K}$ is 1-Lipschitz on the whole domain $\mathbb{R}^d$ and bounded. 
Formally, 
for $f\in 
\Lip_{1, K}$, by \cite[Theorem 1]{minty1970extension}, there exists a function 
$\tilde{f}\in \Lip_{1, \mathbb{R}^d}$ with $\tilde{f}(x) = f(x)$ for all $x \in 
K$ and $\sup_{x\in\mathbb{R}^d} |\tilde{f}(x)| = \sup_{x\in K} |f(x)|$. 
Since 
for the statement of Theorem \ref{thm:main} only the values of $f$ on 
$K$ 
are of interest, one can replace $f$ by $\tilde{f}$ and approximate $\tilde{f}$ 
instead.

Finally, we work with $K = (0, 1)^d$ which 
can be done without loss of generality, the reason being as follows: 
Suppose the statements hold for $K = (0, 1)^d$ and we want to prove them for
general $\tilde{K}$: Take $l_i := \inf\{x_i : x \in \tilde{K}\}$ and 
$u_i := 
\sup\{x_i : x \in \tilde{K}\}$ for $i=1, ..., d$ and set $l=(l_1, ..., l_d)$ 
and $M := 
\max\{ u_i - l_i : i \in \{1, ..., d\}\}$. Take any $\tilde{f} \in 
\Lip_{1, \tilde{K}}$ and $\varepsilon > 0$ and define 
\[f(x) := 
\frac{\tilde{f}\big(M x - l\big)}{M} ~~ \text{ for } x \in \mathbb{R}^d.\] 
Then  
$f \in \Lip_{1, K}$ (where we already used that $\tilde{f}$ is assumed to 
be $1$-Lipschitz on $\mathbb{R}^d$). By approximating $f$ by a 
function $f_m \in \Lip_{1, K}^m$ on $K$ up to accuracy 
$\varepsilon/M$ and setting $\tilde{f}_m(x) := M f_m((x+l)/M)$, we get 
$\tilde{f}_m \in 
\Lip_{1, \tilde{K}}^m$ and the desired approximation of $\tilde{f}$ 
by $\tilde{f}_m$.

\subsection{Proof of Theorem \ref{thm:main}: first part}
\label{subsec:proofpart1}
Fix $f\in \Lip_{1, K}$ and $\varepsilon 
> 0$. We will show that there exists some $m\in \mathbb{N}$ and $f_m \in 
\Lip_{1, K}^m$ such that $\|f - f_m\|_{\infty, K} \leq \varepsilon$.

Define $\hat{f} := (1-\varepsilon/2) f$ and note $\sup_{x\in K} |\hat{f}(x) - 
f(x)| \leq \varepsilon/2$ (where we used the normalization of $\|\cdot\|$) and 
w.l.o.g.~$\hat{f} \in 
\Lip_{1-\varepsilon/2,\mathbb{R}^d}$. 
By \cite[Theorem 
1]{azagra2007smooth} 
there is a 
smooth (i.e., $C^\infty$) function $\tilde{f} \in 
\Lip_{1-\varepsilon/4, \mathbb{R}^d}$ that satisfies $\|\tilde{f} - 
\hat{f}\|_\infty < \varepsilon/4$. Hence also $\|\tilde{f} 
- f\|_{\infty, K}
\leq 3\varepsilon/4$. 

We next approximate $\tilde{f}$ and its 
first partial
derivatives by a function $f_m \in N^m$. The desired accuracy 
depends on the norm $\|\cdot\|$. Since all norms on $\mathbb{R}^d$ are 
equivalent, we can find a constant $C > 0$ such that $\|\cdot\|_1 \leq C 
\|\cdot\|$. Set $\delta := \min\{\varepsilon/4, \varepsilon/(4dC)\}$ 
and find a function $f_m \in N^m$ which satisfies
\begin{align}
\label{eq:approx}
\Big\|\frac{\partial f_m}{\partial x_i} - \frac{\partial 
	\tilde{f}}{\partial 
	x_i}\Big\|_{\infty, K} &\leq \delta \text{ for all } i\in\{1, ..., d\}, \\
\|f_m - \tilde{f}\|_{\infty, K} &\leq \delta.
\end{align}
This can be done by \cite[Theorem 4.1]{pinkus1999approximation} for the case of 
a 
continuously differentiable activation function, and by \cite[Theorem 
4.3]{ito1994differentiable} for the case of the ReLU.\footnote{In case of the 
	ReLU, $\frac{\partial{f_m}}{\partial x_i}$ is understood in the weak sense. 
	To 
	apply 
	\cite[Theorem 4.3]{ito1994differentiable} to the ReLU, note that $G(x) = 
	\max\{0, x\} - 2 \max\{0, x+1\} + \max\{0, x+2\}$ gives the desired linear 
	combination of scaled shifted rotations of the ReLU.}

It then holds \[\|f_m
- f\|_{\infty, K} \leq \|f_m - \tilde{f}\|_{\infty, K} + \|\tilde{f} - 
f\|_{\infty, K} \leq \frac{\varepsilon}{4} + \frac{3\varepsilon}{4} = 
\varepsilon.\]
It remains to show that $f_m \in \Lip_{1, K}$. First, we consider 
the 
case where the activation function is continuously differentiable and hence so 
is $f_m$. We use Lemma \ref{lem:equiv} in the appendix and show that $f_m$ 
satisfies part $(i)$ of the lemma. For $x \in K, v \in 
\mathbb{R}^d$ it holds
\begin{align*}
|Df_m(x) \cdot v| &\leq |D f_m(x) \cdot v - D \tilde{f}(x) \cdot v| + 
|D\tilde{f}(x) \cdot 
v| \\ 
&\leq \Big| \sum_{i=1}^d \Big(\frac{\partial f_m}{\partial x_i}(x) - 
\frac{\partial \tilde{f}}{\partial x_i}(x)\Big) v_i \Big| + 
\big(1-\frac{\varepsilon}{4}\big) \|v\|\\
&\leq \sup_{\hat{x}\in K, i\in\{1, ..., d\}} \Big|\frac{\partial 
	f_m}{\partial 
	x_i}(\hat{x}) - \frac{\partial 
	\tilde{f}}{\partial 
	x_i}(\hat{x})\Big| \,d \,\|v\|_1 + \big(1-\frac{\varepsilon}{4}\big) 
\|v\|\\
&\leq \delta \,d \, C  \, \|v\| + \big(1-\frac{\varepsilon}{4}\big) \|v\| 
\\&\leq \|v\|,
\end{align*}
where we used Lemma \ref{lem:equiv} for $\tilde{f}$. This shows $f_m \in 
\Lip_{1, K}$. 

We now consider the case of the ReLU. We choose a standard mollifier 
$\eta_\kappa$ for $\kappa > 0$.\footnote{See, e.g., \cite{evans2002partial}. 
	The mollifier is taken w.r.t.~the euclidean norm, and $B(x, \kappa)$ 
	denotes 
	the open ball around $x$ of radius $\kappa$ w.r.t.~the euclidean norm.} We 
define 
$f_{m, \kappa} := f_m * \eta_\kappa$ and $\tilde{f}_\kappa := \tilde{f} * 
\eta_\kappa$. Note that there exists $\lambda(\kappa) > 0$ with 
$\lambda(\kappa) \rightarrow 0$ for $\kappa \rightarrow 0$ such that 
$\sup_{i\in\{1, ..., d\}}\sup_{x\in 
	K}\big|\frac{\partial\tilde{f}_\kappa}{\partial 
	x_i}(x)-\frac{\partial\tilde{f}}{\partial x_i}(x)\big| \leq 
\lambda(\kappa)$ and $\|f_m - f_{m, \kappa}\|_{\infty, K} \leq 
\lambda(\kappa)$. 
Further, we note that for $i\in\{1, ..., d\}$ and $x \in K$ it holds
\begin{align*}
\Big|\frac{\partial\tilde{f}_\kappa}{\partial 
	x_i}(x)-\frac{\partial f_{m, \kappa}}{\partial x_i}(x)\Big| &= \Big| 
\int_{B(0, \kappa)}
\Big(\frac{\partial \tilde{f}}{\partial x_i} (x-y) - \frac{\partial 
	f_m}{\partial x_i} (x-y) \Big) \eta_\kappa(y) \,dy \Big|\\ &\leq 
\sup_{\hat{x} \in (-\kappa, 
	1+\kappa)^d}\Big|\frac{\partial\tilde{f}}{\partial 
	x_i}(\hat{x})-\frac{\partial f_{m}}{\partial x_i}(\hat{x})\Big|.
\end{align*}
In the following, we will assume w.l.o.g.~that $\sup_{\hat{x} \in (-\kappa, 
	1+\kappa)^d}\Big|\frac{\partial\tilde{f}}{\partial 
	x_i}(x)-\frac{\partial f_{m}}{\partial x_i}(x)\Big| \leq \delta$ holds for 
all $\kappa < 1$. The reason we can make this assumption without loss of 
generality is that the 
approximations in Equation \eqref{eq:approx} may be taken for $K = 
(-1, 3)^d$, 
since $f$ (and hence $\tilde{f}$) can be assumed to be Lipschitz on 
$\mathbb{R}^d$ as argued 
in Subsection 
\ref{subsec:scaling}.
It 
then holds for $x \in K, v \in \mathbb{R}^d$,
\begin{align*}
&|Df_{m, \kappa}(x) \cdot v|\\ &\leq |D\tilde{f}_\kappa(x) \cdot v - D f_{m, 
	\kappa}(x) \cdot v| + |D \tilde{f}_\kappa(x) \cdot v - D \tilde{f}(x) \cdot 
v| 
+ |D \tilde{f}(x) \cdot v| \\
&\leq d\,\|v\|_1 \Big(\sup_{i\in\{1, ..., d\}} \sup_{\hat{x} \in K}  
\Big|\frac{\partial\tilde{f}_\kappa}{\partial 
	x_i}(\hat{x})-\frac{\partial f_{m, \kappa}}{\partial x_i}(\hat{x})\Big|
+ \sup_{i\in\{1, ..., d\}}\sup_{\hat{x}\in 
	K}\Big|\frac{\partial\tilde{f}_\kappa}{\partial 
	x_i}(\hat{x})-\frac{\partial\tilde{f}}{\partial x_i}(\hat{x})\Big|\Big) 
+ \big(1-\frac{\varepsilon}{4}\big) \|v\| \\
&\leq d\,C\,\|v\| \Big( \sup_{i\in\{1, ..., d\}}\sup_{\hat{x}\in (-\kappa, 
	1+\kappa)^d} 
\Big|\frac{\partial\tilde{f}}{\partial 
	x_i}(\hat{x})-\frac{\partial f_{m}}{\partial x_i}(\hat{x})\Big| + 
\lambda(\kappa)\Big) 
+ \big(1-\frac{\varepsilon}{4}\big) \|v\| \\
&\leq d\,C\,\|v\|\,\delta + d\,C\,\|v\|\,\lambda(\kappa) + 
\big(1-\frac{\varepsilon}{4}\big) 
\|v\| \\
&\leq (1+d\,C\,\lambda(\kappa)) \|v\|
\end{align*}
and hence $f_{m, \kappa}$ is $(1+d\,C\,\lambda(\kappa))$-Lipschitz on $K$ 
according 
to Lemma \ref{lem:equiv}.
Thus, for all $x, y\in K$, we have
\begin{align*}
|f_m(x) - f_m(y)| &\leq |f_m(x) - f_{m, \kappa}(x) + f_{m, \kappa}(x) -f_{m, 
	\kappa}(y) + f_{m, \kappa}(y) - f_m(y)|\\ &\leq 2 \lambda(\kappa) + 
(1+d\,C\,\lambda(\kappa))\|x-y\|,
\end{align*}
and taking $\kappa \rightarrow 0$ yields $f_m \in \Lip_{1, K}$. The first part 
of the proof is complete.

\subsection{Proof of Theorem \ref{thm:main}: second part}
\label{subsec:proofpart2}
We prove that the size $m$ of the networks may be chosen only depending on 
$\varepsilon$, but independently of $f$. We still assume that any Lipschitz 
function satisfies $f(0) = 0$, since shifting neural network functions by 
constants does not affect their size. We choose some compact set 
$\hat{K} \subset \mathbb{R}^d$ 
with $K \subset \hat{K}$. We set $\mathcal{F} := \{ g: \hat{K} \rightarrow 
\mathbb{R} : 
g(0) = 0 \text{ and } g \text{ is $L$-Lipschitz}\}$. Since $\mathcal{F}$ is 
bounded, convex, closed and equicontinuous, by the Arzel\`a-Ascoli theorem 
$\mathcal{F}$ is compact with respect to the uniform norm. 

Hence, for a given $\varepsilon > 0$ we can find $g_1, ..., g_n \in 
\mathcal{F}$ such that \[\sup_{g \in \mathcal{F}} \inf_{i \in\{1, ..., n\}} 
\|g - g_i\|_{\infty, \hat{K}} \leq \frac{\varepsilon}{2}.\] We can 
approximate $g_1, ..., g_n$ (respectively their extensions to the whole domain 
$\mathbb{R}^d$) as in Subsection \ref{subsec:proofpart1} up to accuracy 
$\frac{\varepsilon}{2}$ by functions $g_1^{m_1}, ..., g_n^{m_n}$ and set $m := 
\max\{m_i : i \in\{1, ..., n\}\}$ so that $g_i^{m_i} \in \Lip_{L, K}^m$ for all 
$i=1, ..., n$. Then, for any $f \in \Lip_{L, K}$, choose an 
extension $\tilde{f} \in \Lip_{L, \mathbb{R}^d}$ by \cite[Theorem 
1]{minty1970extension} and set $g := \tilde{f}_{|\hat{K}} \in \mathcal{F}$ to 
be the restriction of $\tilde{f}$ to $\hat{K}$. Choose $i \in \{1, 
..., n\}$ such that $\|g_i - g|_{\infty, \hat{K}} \leq 
\frac{\varepsilon}{2}$ and obtain the desired approximation of $f$ by 
$g_i^{m_i}$. The proof of Theorem \ref{thm:main} is complete. \qed
\appendix

\section{Lipschitz continuity and directional derivatives}
The following lemma is a slight simplification of \cite[Section 5.8, Theorem 
4]{evans2002partial}.
\begin{lemma}
	\label{lem:equiv}
	Let $f : \mathbb{R}^d \rightarrow \mathbb{R}$ be continuously 
	differentiable, fix $L > 0$ and assume that $K$ is open and convex. Then 
	the following 
	are equivalent:
	\begin{itemize}
		\item[(i)] For all $x \in K$ and $v\in \mathbb{R}^d$ it holds $|D f(x) 
		\cdot 
		v| 
		\leq L \|v\|$
		\item[(ii)] $f \in \Lip_{L, K}$
	\end{itemize}
	\begin{proof}
		Assume $(i)$ holds and take $x, y \in K$. Then it holds
		\[
		|f(x)-f(y)| = \Big|\int_0^1 Df(tx + (1-t)y) \cdot (x-y) \,dt\Big| \leq 
		\int_0^1|Df(tx + 
		(1-t)y) \cdot (x-y)| \,dt \leq L\|x-y\|
		\]
		since by convexity $tx+(1-t)y \in K$ for all $t \in (0, 1)$. 
		Thus $f \in \Lip_{L, K}$.
		
		Conversely, assume $(ii)$ holds. For $x \in K, v \in \mathbb{R}^d$ it 
		holds
		\[
		|Df(x)\cdot v| = \Big|\lim_{h\rightarrow 0}\frac{f(x+hv) - 
			f(x)}{h}\Big| \leq \lim_{h\rightarrow 0} \frac{L \|hv\|}{h} = L\|v\|
		\]
		since $x+hv \in K$ for $h$ small enough since $K$ is open. This shows 
		(i).
	\end{proof}
\end{lemma}

\section*{Acknowledgments}
The author thanks Jonas Blessing, Luca De Gennaro Aquino, Marlene Koch and 
Michael Kupper for helpful discussions and remarks on an earlier version of 
this note.

\bibliographystyle{abbrv}
\bibliography{lipbib.bib}

\end{document}